\documentclass{article}

\usepackage{graphicx}
\usepackage[numbers]{natbib}
\setcitestyle{numbers} 

\usepackage{color}
\usepackage[ruled,vlined]{algorithm2e}
\usepackage{amsmath}

\usepackage{fullwidth}
\usepackage{wrapfig}

\usepackage[final]{nips_2018}

\usepackage[utf8]{inputenc} 
\usepackage[T1]{fontenc}    
\usepackage{url}            
\usepackage{booktabs}       
\usepackage{amsfonts}       
\usepackage{nicefrac}       
\usepackage{microtype}      

\title{Bridging the Generalization Gap: Training Robust Models on Confounded Biological Data}

\author{
  Tzu-Yu ~Liu\thanks{: authors with equal contributions.
 Please direct questions to authors@freenome.com.}, ~ Ajay ~Kannan$^{*}$, ~ Adam ~Drake, ~ Marvin ~Bertin, ~ Nathan ~Wan\\
  Freenome Inc.\\
  South San Francisco, CA \\
  \texttt{\{joyce.liu, ajay.kannan, adam.drake, marvin.bertin, nwan\}@freenome.com} \\
}

\begin{document}

\maketitle

\begin{abstract}
  Statistical learning on biological data can be challenging due to confounding variables in sample collection and processing. Confounders can cause models to generalize poorly and result in inaccurate prediction performance metrics if models are not validated thoroughly. In this paper, we propose methods to control for confounding factors and further improve prediction performance. We introduce OrthoNormal basis construction In cOnfounding factor Normalization (ONION) to remove confounding covariates and use the Domain-Adversarial Neural Network (DANN) to penalize models for encoding confounder information. We apply the proposed methods to simulated and empirical patient data and show significant improvements in generalization.
\end{abstract}

\section{Introduction}
 Confounding is most likely to occur when incomplete understanding of the data generation process renders factoring out confounding signal difficult. Biological datasets derived from patient samples are prone to confounding due to artifactual differences in sample collection \cite{libbrecht2015machine}. Consider a binary classification problem, 
 where positive and negative samples are sourced from hospital A and B, respectively. It becomes unclear if a classifier is predicting class labels or the source hospital. Stratified sampling to control for confounding variables is often infeasible due to sample procurement difficulties, lab processing batch effects, and incomplete knowledge of potential confounders.  As \citeauthor{leek2010tackling} pointed out, these often overlooked batch effects could result in incorrect conclusions \citep{leek2010tackling}. A recent study also highlighted that performance metrics in cancer detection using sequencing data can be inaccurate when using standard k-fold cross-validation due to sequencing batch effects and sample sourcing biases \citep{crc_paper}. Confounding is a known problem \cite{kulasingam2008strategies, chechlinska2010systemic}, but it is often difficult to detect or correct, and requires careful consideration when applying machine learning methods \cite{Chuang2018}.
 
 Since stratifying samples by confounders is often infeasible, correcting the resulting biases in model inputs and representations is an important area of open research.  There has been previous work on bias correction in sequencing data, including Hidden Covariates with Prior \citep{hcp} and ComBat \citep{combat}, which use covariate labels to normalize sequencing data. However, these methods require test set covariate labels and model re-fitting at test time, which make freezing and presenting a model for regulatory review and validation on previously unseen populations challenging. There are also domain knowledge-based sequencing technical bias correction methods, such as LOWESS GC bias correction \citep{lowess, lowess_gc}. Our methods perform equivalently without relying on domain knowledge, and can be applied to generic machine learning problems. 
 
 To avoid models learning confounders, we build transformations such that samples are less confounded in the transformed space. Our goal is to find a function $f$ such that the image $f(X)$ enables robust learning, where $X$ represents the biological data. Note that the confounding covariates may be used to learn $f$, but are not arguments of $f$. To this goal, we (1) develop a method named OrthoNormal basis construction In cOnfounding factor Normalization (ONION) for factoring out confounding covariates and (2) apply the Domain-Adversarial Neural Network (DANN) \cite{ganin2016domain} to refrain from learning confounders. The first approach constructs a basis such that some basis vectors span the confounded vector space and the remaining basis vectors span the non-confounded complement vector space. 
 The second method seeks latent spaces that accurately encode the target label and poorly encode confounders. We conduct experiments on both simulated data and real world clinical data to show that ONION and DANN can effectively bridge generalization gaps caused by confounders.

\section{Methods}

Let $X$: $n \times p$ be the observed data, where $n$ is the number of observations and $p$ is the number of features. The underlying data generation mechanism involves several factors including the disease status and possibly several confounders. Let $Y_1, Y_2, ..., Y_{k-1}$ represent the $k-1$ confounders, where $Y_i$ is $n \times 1$, $i \in \{ {1,2,...,k-1} \}$, e.g., the age, sex, sample source institution, etc., and let $Y_k$ : $n \times 1$ represent the label of interest, e.g., the clinical disease labels.

\subsection{Orthonormal basis construction in confounding factor normalization (ONION)}
Assume that $X$ has been centered. The objective of ONION is to rewrite $X$ as $X=X_c+X_n$, in which $X_c$ is associated with $Y_i$'s for $i=1,2,..,k-1$, representing the confounder signal, and $X_n$ is the residual after factoring out covariates, i.e., the normalized data. This is equivalent to finding an orthonormal basis $W$ for ${\bf R}^p$ such that when projecting $X$ onto these vector spaces, the association with the confounders can be deconvolved.
\[
W=\begin{bmatrix}
|&|&& |\\
w_1&w_2&...& w_p\\
|&|& … & |\\
\end{bmatrix}, \ WW^T=I, \ where \ w_i \in {\bf R}^p \  denotes \ the \ i^{th} \ basis \ vector. \]\[
X=XWW^T=X_c + X_n, \ where \ X_c=\sum_{i<k} Xw_iw_i^T \ and \ X_n=\sum_{i\geq k} Xw_iw_i^T.
\]

There are many methods in the literature to construct $W$ and factorize $X$. For example, principle component analysis (PCA) sequentially identifies these basis vectors to maximize the variance \cite{friedman2001elements}, i.e., $w_1=argmax_{||w||=1} w^TX^TXw$.  However, PCA is unsupervised, and top variations may not be associated with the response. Partial least squares (PLS) and canonical correlation analysis (CCA) are supervised dimension reduction methods and the latent variables can be used as predictors, often giving better performance than standard regression in large $p$ small $n$ scenarios \cite{friedman2001elements,de1993simpls, hotelling1936relations}. One can adopt the objective functions in PLS and CCA, but change the orthogonality conditions to form an orthonormal basis to factor out the confounders. The orthonormal constraints allow us to represent data in terms of the basis vectors simply by taking the product as projection. Here we show the modified optimization motivated by PLS:
\begin{equation}
\label{onion_optimization}
w_i = argmax_w \ w^TX^TY_iY_i^TXw, \ s.t. \ ||w||=1 \ and \ w^Tw_j=0 \ for \ j<i<k.
\end{equation}
Notice that the term $w^TX^TY_iY_i^TXw$ finds weight vector that maximizes the covariance between the latent variables $Xw$ and confounder $Y_i$. The remaining weight vectors $w_k, w_{k+1}, ..., w_p$ are the basis vectors for the null space of $\{ w_1, w_2, ..., w_{k-1} \}$. Since we are interested in building models to predict $Y_k$ but not the confounders $Y_1, Y_2, ..., Y_{k-1}$, we train models on $X_n$ instead of on $X$. A sequential algorithm using power iteration \cite{moon2000mathematical} and applying 
deflation to satisfy the orthogonality condition in the optimization problem (\ref{onion_optimization}) is presented in  Algorithm \ref{onion_algo}. ONION peels away layers of confounders’ effects sequentially, hence the acronym.

\begin{wrapfigure}[18]{R}{0.4\textwidth}
\begin{algorithm}[H]
\label{onion_algo}
\small
\SetAlgoLined
 initialization $W= [\ \ ]$;
 (an empty matrix)\;
 $X_d=X$\;
\For{$i \gets 1$ to $k-1$} {   
    \uIf{$i>1$}{
        $X_d=X_d-X^Tw_{i-1}w_{i-1}^T$\;}
    Randomly initialize $u_0$ \;
    $\tau=0$ \;
    \While{stopping criterion is not satisfied}{
    $\tau=\tau+1$\;
    $u_{\tau}=X_d^TY_iY_i^TX_du_{\tau-1}$ \;
    $u_{\tau}=u_{\tau}/||u_{\tau}||$\;
    }
    $W=[W;u_{\tau}]$; (append $u_{\tau}$ to $W$) 
}
$X_n=X-XWW^T$
\caption{ONION}
\end{algorithm}
\end{wrapfigure}

\subsection{Domain-adversarial neural network}
Ganin et al.'s Domain-Adversarial Neural Network (DANN) is a feed-forward neural network that shares at least one hidden layer between a target prediction network and a confounder prediction network \cite{ganin2016domain}.  Gradients are reversed between the confounder prediction network and the shared layers to remove confounding signal from model representations. See Figure 1 from \cite{ganin2016domain} for a network diagram.  In our use case, we train $k-1$ networks $f_{Y_i}(g(X)), i=1, ..., k-1$ to predict the $k-1$ confounders and $f_{Y_k}(g(X))$ to predict the clinical label. $g$, the shared feature extractor, can be any differential function of $X$. Training proceeds by alternating stochastic gradient descent updates to:
\begin{enumerate}
\item $\theta_{g_s}$ and $\theta_{k_s}$, i.e., $g$'s and $f_{Y_k}$'s parameters at step $s$, according to the rules $\theta_{g_{s+1}} = \theta_{g_s} - \alpha\frac{dL_k}{d\theta_{g_s}}$ and $\theta_{k_{s+1}} = \theta_{k_s} - \alpha\frac{dL_k}{d\theta_{k_s}}$, where $L_k$ is the cross entropy loss between predicted and actual clinical labels $Y_k$.
\item $\theta_{g_s}$ and $\theta_{i_s}$, for $i=1, ..., k-1$, i.e.,  $g$'s and $f_{Y_i}$'s parameters at step $s$, according to the rules $\theta_{g_{s+1}} = \theta_{g_s} + \alpha \sum_{i=1}^{k-1}{ \frac{dL_i}{d\theta_{g_s}}}$ and $\theta_{i_{s+1}} = \theta_{i_s} - \alpha\frac{dL_i}{d\theta_{i_s}}$, where $L_i$ is the loss between predicted and actual confounder $Y_i$.
\end{enumerate}

\section{Experiments}

We present results on three datasets to showcase improvements in generalization by correcting for confounders using ONION and DANN.  In each scenario, we compare (a) logistic regression (logreg) with and without ONION and (b) an MLP and DANN with the same clinical label prediction architecture using the same train and test data per fold across models.

\subsection{Confounded data simulation}

We first study confounding in a well-understood setting by simulating confounded data (see supplementary materials). The left panel in Figure \ref{simulation_results} shows the distribution of the confounded samples in the first fold of the 5-fold cross validation when $n=6000$. Notice that the two classes can be perfectly separated by the confounder value in the training set, but not in the test set. Hence, a classifier that learns to predict the confounder will not be robust to confounder value changes in the test set. We compare ONION and DANN against ANalysis of COVAriance (ANCOVA), a univariate feature selection method to filter confounded variables \cite{huitema2011analysis}. Figure \ref{simulation_results}
shows the classification performance in terms of the Area Under the receiver operating characteristic Curve (AUC) as the number of simulated samples varies. The performance gaps between logreg with and without ONION and multilayer perceptron (MLP) versus DANN become apparent as the sample size increases. ANCOVA improves the logreg performance marginally, and becomes impractical as the number of features $p$ increases, so we focus on ONION and DANN in subsequent experiments.

\begin{figure}[t!]
    \centering
        \includegraphics[width=0.8\linewidth]{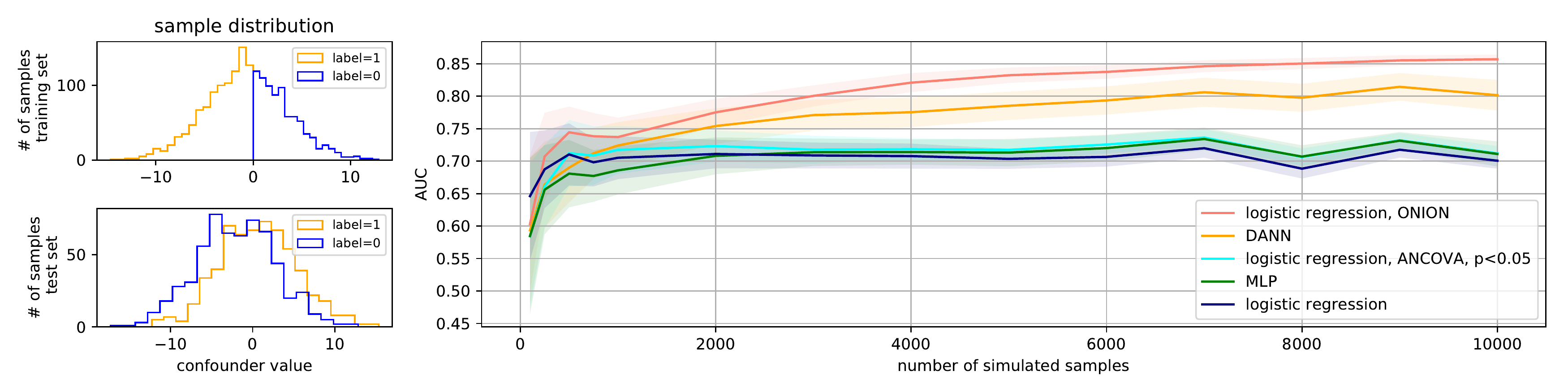}
    \caption{ \footnotesize Comparison of no confounder correction versus with correction on balanced test sets using simulated data.  The graphs on the left show the distribution of samples with respect to the confounder and labels in the training and test sets when $n=6000$.  The graph on the right compares uncorrected methods to corrective methods as dataset size increases. The curves represent the mean AUC over 50 trials, and the shaded area represents $+/-$ standard error. The method logreg ANCOVA filters features with insignificant association to the label of interest (p-value $\geq$ 0.05) after removing univariate confounder effects. \label{simulation_results}}
\end{figure}

\subsection{Sequencing data from clinical cancer samples}

In both of the following scenarios, we are tasked with predicting whether a patient has cancer or not after sequencing cell-free DNA from the patient's blood \cite{snyder2016cell,adalsteinsson2017scalable}.  We featurize the sequencer reads by counting the number of DNA fragments that align to a set of non-overlapping 50 kilobase bins of the human genome, which provides us with $61{,}775$ features. Since we tend to control for known confounders in our clinical datasets, we artificially introduce confounding to study its impact on generalization using methods described in the supplementary materials. The dataset consists of 520 cancer patients and 214 healthy patients. Experiments are conducted with 5-fold cross validation. 

\subsubsection{Confounding by biological sex}

An example of sample collection bias is when most cancer samples are sourced from a particular population subgroup and most healthy samples are sourced from outside that subgroup.  We construct a training set where cancer patients are largely female and healthy patients are largely male, whereas the test set is more balanced (see the sample distribution in the supplementary materials). As shown in Table~\ref{confounder-gender-gc}, without correcting for confounders (logreg and MLP), the performance is artificially high when the test set is confounded in the same manner as the training set, as is the case when performing k-fold cross validation.  However, on the entire test set, without subsampling to mimic training set confounding, uncorrected models perform far worse than corrected models (logreg with ONION and DANN). The reference in Table~\ref{confounder-gender-gc} represents the result using logreg with sex chromosomes removed, which serves as the benchmark if prior knowledge of the features is given. 

\subsubsection{Confounding by GC bias}
Sequencing data often exhibits GC bias, a sample-level technical bias in over- and under-representation of DNA fragments with varying amounts of G and C bases.  We construct a scenario in which labels in our training set are confounded by the level of GC bias (see supplementary materials for more detail). 
In Table~\ref{confounder-gender-gc}, we see that GC confounding is more subtle than sex confounding, though the same trends of overinflated metrics without confounder correction and improved performance on unconfounded test sets with confounder correction hold.

\section{Discussion and conclusion}

In this paper, we propose a novel method, ONION, and apply DANN to correct for confounding factors. Simulated experiments show that both methods outperform univariate ANCOVA screening. Our experiments using clinical cancer data also show that ONION and DANN generalize well, reducing the gap between the performance on the entire test set and a confounded subsampled test set.  Experimenting with confounding by biological sex is particularly informative because an unconfounded model should not build predictions upon inferring sex using the sex chromosomes, chrX and chrY. We indeed notice that the logistic regression model without ONION has opposite signs on chrX and chrY weights (see supplementary materials), whereas using ONION removes this trend. A direction of future work is to incorporate generative networks and data augmentation into our models \cite{goodfellow2014generative, bonn2018realistic,zhang2017mixup} and determine DANN's ability to remove nonlinear confounding effects of varying degrees.  Without confounder correction methods, the gap between the performance on the entire test set and the confounded test in our simulated and clinical datasets set points out the ease of mistakenly presenting inaccurate performance in the machine learning community. We hope this work not only equips researchers with new methods to control for confounding, but also spurs additional work to detect and correct for confounding in biological datasets.
\begin{table}
\small
  \caption{\footnotesize Confounding experiments using clinical cancer data. In the sex confounding experiment, reference represents logreg trained without sex chromosomes. In the GC confounding experiment, reference represents logreg trained after LOWESS GC correction. These references serve as benchmark solutions when prior domain knowledge is given. The confounded test set is the maximally sized subset of the entire test set with the same level of confounding as the training set, which represents using k-fold cross validation on confounded data.  ONION and DANN are able to correct for confounding and often achieve similar performance to the reference.}
  \label{confounder-gender-gc}
  \centering
  \begin{tabular}{  lccccc}
    \toprule 
    & \multicolumn{2}{c}{Sex: mean AUC (SD)} 
    & \multicolumn{2}{c}{GC: mean AUC (SD) }\\
    \cmidrule(r){2-3} 
    \cmidrule(r){4-5}
    method     & entire test set  & confounded test set   &  entire  test set & confounded test set \\
    \midrule
    logreg & 0.59 (0.04)   &   1.00 (0.00) &  0.82 (0.08) & 0.98 (0.01) \\
    logreg, ONION  & 0.91 (0.01) &  0.68 (0.06)  & 0.93 (0.02)   & 0.88 (0.07) \\
    MLP & 0.61 (0.06)   & 0.79 (0.25) & 0.82 (0.11) & 0.93 (0.09) \\
    DANN & 0.78 (0.07) & 0.80 (0.08)  & 0.86 (0.05) & 0.93 (0.07)  \\
    reference & 0.92 (0.02)  & 0.95 (0.01) & 0.92 (0.02) & 0.86 (0.04) \\
    \bottomrule 
  \end{tabular}
\end{table}

\subsubsection*{Acknowledgments}
The authors gratefully acknowledge Lena Cheng, Riley Ennis, Signe Fransen, Girish Putcha and David Weinberg (in alphabetical order) for their extensive suggestions, feedback,  and editorial support. 

\medskip

{\small
\bibliography{nips_2018}}

\begin{thebibliography}{24}
\providecommand{\natexlab}[1]{#1}
\providecommand{\url}[1]{\texttt{#1}}
\expandafter\ifx\csname urlstyle\endcsname\relax
  \providecommand{\doi}[1]{doi: #1}\else
  \providecommand{\doi}{doi: \begingroup \urlstyle{rm}\Url}\fi

\bibitem[Libbrecht and Noble(2015)]{libbrecht2015machine}
Maxwell~W Libbrecht and William~Stafford Noble.
\newblock Machine learning applications in genetics and genomics.
\newblock \emph{Nature Reviews Genetics}, 16\penalty0 (6):\penalty0 321, 2015.

\bibitem[Leek et~al.(2010)Leek, Scharpf, Bravo, Simcha, Langmead, Johnson,
  Geman, Baggerly, and Irizarry]{leek2010tackling}
Jeffrey~T Leek, Robert~B Scharpf, H{\'e}ctor~Corrada Bravo, David Simcha,
  Benjamin Langmead, W~Evan Johnson, Donald Geman, Keith Baggerly, and Rafael~A
  Irizarry.
\newblock Tackling the widespread and critical impact of batch effects in
  high-throughput data.
\newblock \emph{Nature Reviews Genetics}, 11\penalty0 (10):\penalty0 733, 2010.

\bibitem[Wan et~al.(2018)Wan, Weinberg, Liu, Niehaus, Delubac, Kannan, White,
  Ariazi, Bailey, Bertin, Boley, Bowen, Cregg, Drake, Ennis, Fransen, Gafni,
  Hansen, Liu, Otte, Pecson, Rice, Sanderson, Sharma, St.~John, Tang, Tzou,
  Young, Putcha, and Haque]{crc_paper}
Nathan Wan, David Weinberg, Tzu-yu Liu, Katherine Niehaus, Daniel Delubac, Ajay
  Kannan, Brandon White, Eric Ariazi, Mitch Bailey, Marvin Bertin, Nathan
  Boley, Derek Bowen, James Cregg, Adam Drake, Riley Ennis, Signe Fransen, Erik
  Gafni, Loren Hansen, Yaping Liu, Gabriel~L Otte, Jennifer Pecson, Brandon
  Rice, Gabriel~E Sanderson, Aarushi Sharma, John St.~John, Catherina Tang,
  Abraham Tzou, Leilani Young, Girish Putcha, and Imran~S Haque.
\newblock Machine learning enables detection of early-stage colorectal cancer
  by whole-genome sequencing of plasma cell-free dna.
\newblock \emph{bioRxiv}, 2018.
\newblock \doi{10.1101/478065}.
\newblock URL \url{https://www.biorxiv.org/content/early/2018/11/24/478065}.

\bibitem[Kulasingam and Diamandis(2008)]{kulasingam2008strategies}
Vathany Kulasingam and Eleftherios~P Diamandis.
\newblock Strategies for discovering novel cancer biomarkers through
  utilization of emerging technologies.
\newblock \emph{Nature Reviews Clinical Oncology}, 5\penalty0 (10):\penalty0
  588, 2008.

\bibitem[Chechlinska et~al.(2010)Chechlinska, Kowalewska, and
  Nowak]{chechlinska2010systemic}
Magdalena Chechlinska, Magdalena Kowalewska, and Radoslawa Nowak.
\newblock Systemic inflammation as a confounding factor in cancer biomarker
  discovery and validation.
\newblock \emph{Nature Reviews Cancer}, 10\penalty0 (1):\penalty0 2, 2010.

\bibitem[Chuang and Keiser(2018)]{Chuang2018}
Kangway~V. Chuang and Michael~J. Keiser.
\newblock Adversarial controls for scientific machine learning.
\newblock \emph{ACS Chemical Biology}, 13\penalty0 (10):\penalty0 2819--2821,
  2018.
\newblock \doi{10.1021/acschembio.8b00881}.
\newblock URL \url{https://doi.org/10.1021/acschembio.8b00881}.

\bibitem[Mostafavi et~al.()Mostafavi, Battle, Zhu, Urban, Levinson, Montgomery,
  and Koller]{hcp}
Sara Mostafavi, Alexis Battle, Xiaowei Zhu, Alexander~E. Urban, Douglas
  Levinson, Stephen~B. Montgomery, and Daphne Koller.
\newblock Normalizing rna-sequencing data by modeling hidden covariates with
  prior knowledge.
\newblock \emph{PLoS ONE}, 8\penalty0 (7):\penalty0 e68141.

\bibitem[Johnson et~al.(2007)Johnson, Li, and Rabinovic]{combat}
W.~Evan Johnson, Cheng Li, and Ariel Rabinovic.
\newblock Adjusting batch effects in microarray expression data using empirical
  bayes methods.
\newblock \emph{Biostatistics}, 8\penalty0 (1):\penalty0 118–127, 2007.

\bibitem[Cleveland(1979)]{lowess}
William~S. Cleveland.
\newblock Robust locally weighted regression and smoothing scatterplots.
\newblock \emph{Journal of the American Statistical Association}, 74\penalty0
  (368):\penalty0 829--836, 1979.

\bibitem[Benjamini and Speed(2012)]{lowess_gc}
Yuval Benjamini and Terence~P. Speed.
\newblock Summarizing and correcting the gc content bias in high-throughput
  sequencing.
\newblock \emph{Nucleic Acids Research}, 40\penalty0 (10):\penalty0 e72, 2012.

\bibitem[Ganin et~al.(2016)Ganin, Ustinova, Ajakan, Germain, Larochelle,
  Laviolette, Marchand, and Lempitsky]{ganin2016domain}
Yaroslav Ganin, Evgeniya Ustinova, Hana Ajakan, Pascal Germain, Hugo
  Larochelle, Fran{\c{c}}ois Laviolette, Mario Marchand, and Victor Lempitsky.
\newblock Domain-adversarial training of neural networks.
\newblock \emph{The Journal of Machine Learning Research}, 17\penalty0
  (1):\penalty0 2096--2030, 2016.

\bibitem[Friedman et~al.(2001)Friedman, Hastie, and
  Tibshirani]{friedman2001elements}
Jerome Friedman, Trevor Hastie, and Robert Tibshirani.
\newblock \emph{The elements of statistical learning}, volume~1.
\newblock Springer series in statistics New York, NY, USA, 2001.

\bibitem[de~Jong(1993)]{de1993simpls}
Sijmen de~Jong.
\newblock {S}{I}{M}{P}{L}{S}: an alternative approach to partial least squares
  regression.
\newblock \emph{Chemometrics and intelligent laboratory systems}, 18\penalty0
  (3):\penalty0 251--263, 1993.

\bibitem[Hotelling(1936)]{hotelling1936relations}
Harold Hotelling.
\newblock Relations between two sets of variates.
\newblock \emph{Biometrika}, 28\penalty0 (3/4):\penalty0 321--377, 1936.

\bibitem[Moon and Stirling(2000)]{moon2000mathematical}
Todd~K Moon and Wynn~C Stirling.
\newblock \emph{Mathematical methods and algorithms for signal processing},
  volume~1.
\newblock Prentice hall Upper Saddle River, NJ, 2000.

\bibitem[Huitema(2011)]{huitema2011analysis}
Bradley Huitema.
\newblock \emph{The analysis of covariance and alternatives: Statistical
  methods for experiments, quasi-experiments, and single-case studies}, volume
  608.
\newblock John Wiley \& Sons, 2011.

\bibitem[Snyder et~al.(2016)Snyder, Kircher, Hill, Daza, and
  Shendure]{snyder2016cell}
Matthew~W Snyder, Martin Kircher, Andrew~J Hill, Riza~M Daza, and Jay Shendure.
\newblock Cell-free dna comprises an in vivo nucleosome footprint that informs
  its tissues-of-origin.
\newblock \emph{Cell}, 164\penalty0 (1):\penalty0 57--68, 2016.

\bibitem[Adalsteinsson et~al.(2017)Adalsteinsson, Ha, Freeman, Choudhury,
  Stover, Parsons, Gydush, Reed, Rotem, Rhoades,
  et~al.]{adalsteinsson2017scalable}
Viktor~A Adalsteinsson, Gavin Ha, Samuel~S Freeman, Atish~D Choudhury, Daniel~G
  Stover, Heather~A Parsons, Gregory Gydush, Sarah~C Reed, Denisse Rotem,
  Justin Rhoades, et~al.
\newblock Scalable whole-exome sequencing of cell-free dna reveals high
  concordance with metastatic tumors.
\newblock \emph{Nature communications}, 8\penalty0 (1):\penalty0 1324, 2017.

\bibitem[Goodfellow et~al.(2014)Goodfellow, Pouget-Abadie, Mirza, Xu,
  Warde-Farley, Ozair, Courville, and Bengio]{goodfellow2014generative}
Ian Goodfellow, Jean Pouget-Abadie, Mehdi Mirza, Bing Xu, David Warde-Farley,
  Sherjil Ozair, Aaron Courville, and Yoshua Bengio.
\newblock Generative adversarial nets.
\newblock In \emph{Advances in neural information processing systems}, pages
  2672--2680, 2014.

\bibitem[Bonn et~al.(2018)Bonn, Machart, Marouf, Magruder, Bansal, Kilian, and
  Krebs]{bonn2018realistic}
Stefan Bonn, Pierre Machart, Mohamed Marouf, Daniel~Sumner Magruder, Vikas
  Bansal, Christoph Kilian, and Christian~F Krebs.
\newblock Realistic in silico generation and augmentation of single cell
  rna-seq data using generative adversarial neural networks.
\newblock \emph{bioRxiv}, page 390153, 2018.

\bibitem[Zhang et~al.(2017)Zhang, Cisse, Dauphin, and
  Lopez-Paz]{zhang2017mixup}
Hongyi Zhang, Moustapha Cisse, Yann~N Dauphin, and David Lopez-Paz.
\newblock mixup: Beyond empirical risk minimization.
\newblock \emph{arXiv preprint arXiv:1710.09412}, 2017.

\bibitem[Browne(1979)]{browne1979maximum}
Michael~W Browne.
\newblock The maximum-likelihood solution in inter-battery factor analysis.
\newblock \emph{British Journal of Mathematical and Statistical Psychology},
  32\penalty0 (1):\penalty0 75--86, 1979.

\bibitem[Klami et~al.(2013)Klami, Virtanen, and Kaski]{klami2013bayesian}
Arto Klami, Seppo Virtanen, and Samuel Kaski.
\newblock Bayesian canonical correlation analysis.
\newblock \emph{Journal of Machine Learning Research}, 14\penalty0
  (Apr):\penalty0 965--1003, 2013.

\bibitem[{Broad Institute}((Accessed: {2018; version
  2.13.2}))]{Picard2018toolkit}
{Broad Institute}.
\newblock Picard tools.
\newblock \url{http://broadinstitute.github.io/picard/}, (Accessed: {2018;
  version 2.13.2}).

\end{thebibliography}

\clearpage

\section{Supplementary Materials}

\subsection{Clinical cancer dataset}

The clinical samples were sourced from various academic institutions and sample collection companies from retrospective studies.  These samples include colorectal cancer samples, the majority of which are early stage cancer samples, and healthy samples as controls.

\begin{center}
\begin{tabular}{ llll }
\hline
~ & Healthy & Cancer & Total \\ 
\hline
Female & 156 & 245 & 401 (55\%) \\ 
\hline
Male & 58 & 275 & 333 (45\%) \\ 
\hline
Total & 214 (29\%) & 520 (71\%) & 734 \\
\hline
\end{tabular}
\end{center}

We preprocess the 50 kilobase bin fragment counts by multiplying each sample $i$'s counts, denoted $c_i$, by $\frac{k}{sum(c_i)}$ ($k$ is a constant) to simulate an equal number of reads per sample.  We then clip and standardize the counts based on the training set, and apply the learned mean, standard deviation, and outlier thresholds to the test set.  We clip outlier counts to the 99th percentile count per feature.

\subsection{Model hyperparameters}
Due to the small number of samples, we restrict the size of the MLP and DANN architectures.  In clinical cancer experiments, we use the first 200 principle components of PCA to reduce input dimensionality and use one hidden layer of 20 units.  In DANN, we use a hidden layer of 20 units to predict the confounder of interest that uses the 20 unit hidden layer from the clinical label prediction network as input.  In all comparisons, we fix hyperparameters for fair comparison.  We train models using cross entropy loss for categorical variables and mean squared error for continuous variables.  We choose the best performing model based on a validation set, using accuracy as the model metric for MLPs and using $L_{k} - \sum_{i=1}^{k-1}{\alpha_i L_{i}}$ (a weighted sum of the clinical label prediction loss and the negative confounder prediction losses) for DANNs.  Models are trained for 6,000 iterations with a batch size of 64 for the clinical label prediction network using Adam and a learning rate of 0.005.  DANN's confounder prediction network is trained three steps for every clinical label prediction update.  Everything is kept the same for the simulated data task, except the hidden layer in both the MLP and DANN is 5 units rather than 20.

\subsection{Confounding procedure}
\subsubsection{Simulated confounding}
We extend the inter-battery factor analysis model studied in \cite{browne1979maximum,klami2013bayesian} to formulate a simulation model for generating confounded data.
\begin{equation}
\label{simulation_model}
\begin{split}
& Z_i  \sim  N(0,I_d), \ i=1,2,...,k \\
& W_{x_i}: \ d \times p, \ W_{x_i}(i,j) \sim N(0,1),\ i=1,2,...,k\\
& W_{y_i} : \ d \times 1, \ W_{y_i}(i) \sim N(0,1), \ i=1,2,...,k\\
& \mathcal{E}_{X} \sim N(0,\sigma^2 I_p)\\
& \mathcal{E}_{y_i} \sim N(0,\sigma^2), \ i=1,2,...,k\\
& \alpha =\big[\alpha_1, \alpha_2, ..., \alpha_k  \big] \sim Dir(\big[s_1, s_2, ..., s_k  \big])\\
& X = \sum_{i=1}^kZ_iW_{x_i}+\mathcal{E}_{X}\\ 
& Y_i = Z_iW_{y_i}+\mathcal{E}_{Y_i}, \ i=1,2,...,k-1\\
& Y_k =  {\bf 1}{\{ \sum_{i=1}^{k-1}\alpha_iY_i+\alpha_kZ_kW_{y_k}+\mathcal{E}_{y_k} >0\}},\\
\end{split}
\end{equation}
in which ${\bf 1}$ is the indicator function, $N(\mu,\Sigma)$ represents the normal distribution with mean $\mu$ and covariance $\Sigma$, and $Dir(\big[s_1, s_2, ..., s_k  \big])$ represents the Dirichlet distribution with concentration parameters $\big[s_1, s_2, ..., s_k  \big]$. $Z_i$'s are the shared latent variables between the independent and dependent variables, and $W_{x_i}$ and $W_{y_i}$ are used for constructing linear combinations of the latent variables to synthesize $X_i$ and $Y_i$, with additive noise $\mathcal{E}_{X}$ and $\mathcal{E}_{y_i}$. The vector $\alpha$ sampled from the Dirichlet distribution controls for the confounding level, i.e., the larger the $\alpha_j$, the higher impact the confounder $Y_j$ on the true label of interest, $Y_k$. 

We set $d=20$, $p=300$, $\sigma=2$, $k=2$ and $s_1=40$, $s_2=50$ in our experiments. To synthesize $n$ samples for the experiment, we first sample $W_{x_i}$, $W_{y_i}$ and $\alpha$ once, and repeatedly sample $Z_i$'s for $n$ times to construct $X$ and $Y_i$'s. We then filter the samples to create skewed distribution in the training set by only allowing samples that have $Y_1<0$ and $Y_2=1$ or $Y_1\geq 0$ and $Y_2=0$, while the test set is kept balanced. 

\subsubsection{Confounding by sex}
\begin{figure}[t!]
    \centering
        \includegraphics[width=0.6\linewidth]{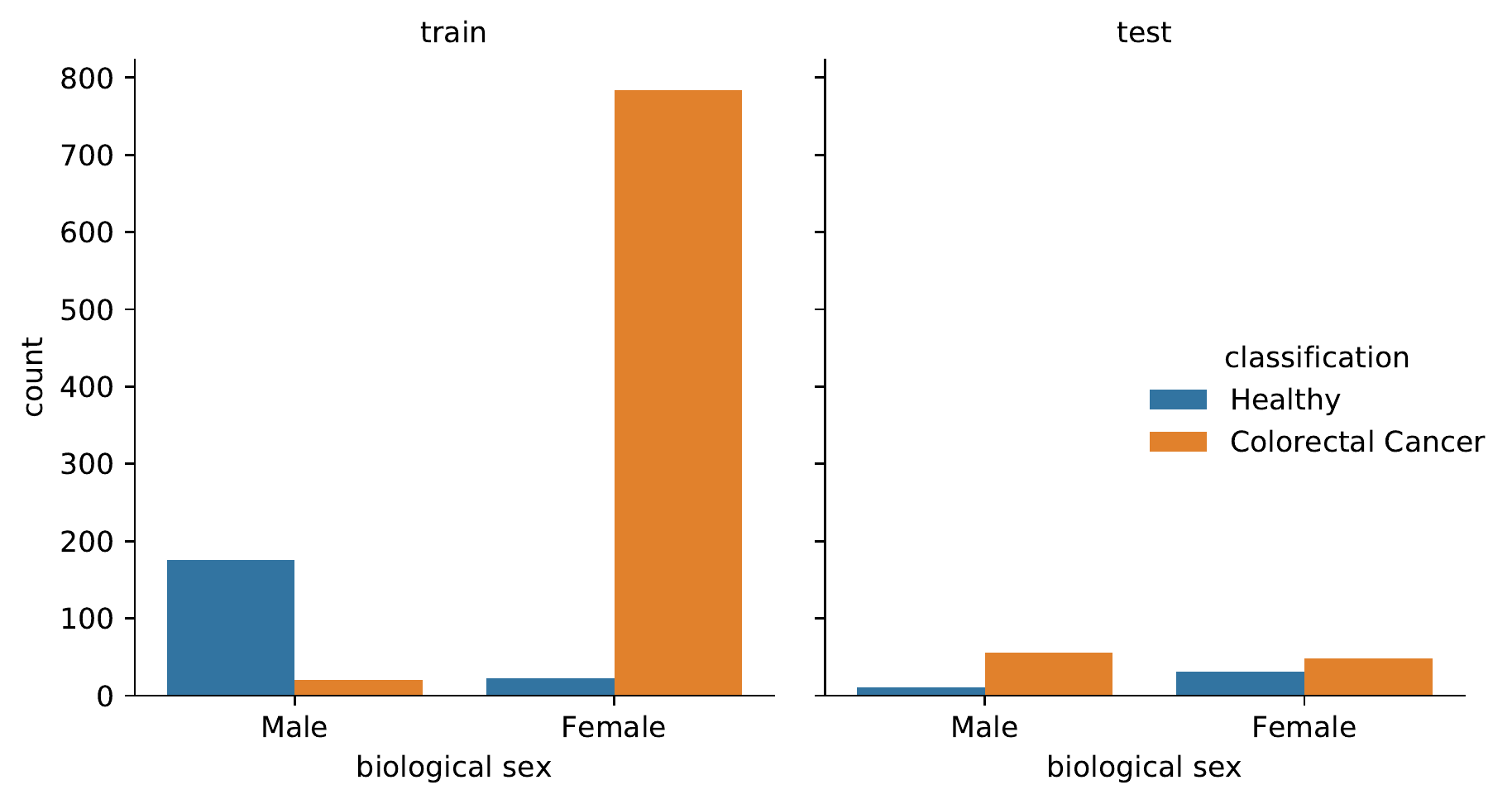}
    \caption{Sample distribution in the experiments of sex confounding. \label{gender_distribution}}
\end{figure}

We construct a training set where cancer patients are largely female and healthy patients are largely male, whereas the test set is balanced. The training set is obtained by randomly removing the male cancer patients and female healthy patients with probability 0.9. Figure \ref{gender_distribution} shows the sample distributions in one of the 5 folds. 

\subsubsection{Confounding by GC bias}
Genomes consist of a heterogeneous mix of chemicals called nucleotides with varying electrochemical properties. These chemicals are guanosine (G), cytosine (C), adenosine (A), and thymidine (T). Guanosine and cytosine have the potential to form three hydrogen bonds while adenosine and thymidine make two. This difference in bonding results in different behavior of DNA fragments based on their nucleotide composition. In particular, the processes that convert DNA molecules into modern DNA sequencer output introduce biases related to the GC-content of the input molecules, which is known as GC bias.

\begin{figure}[t!]
    \centering
        \includegraphics[width=0.6\linewidth]{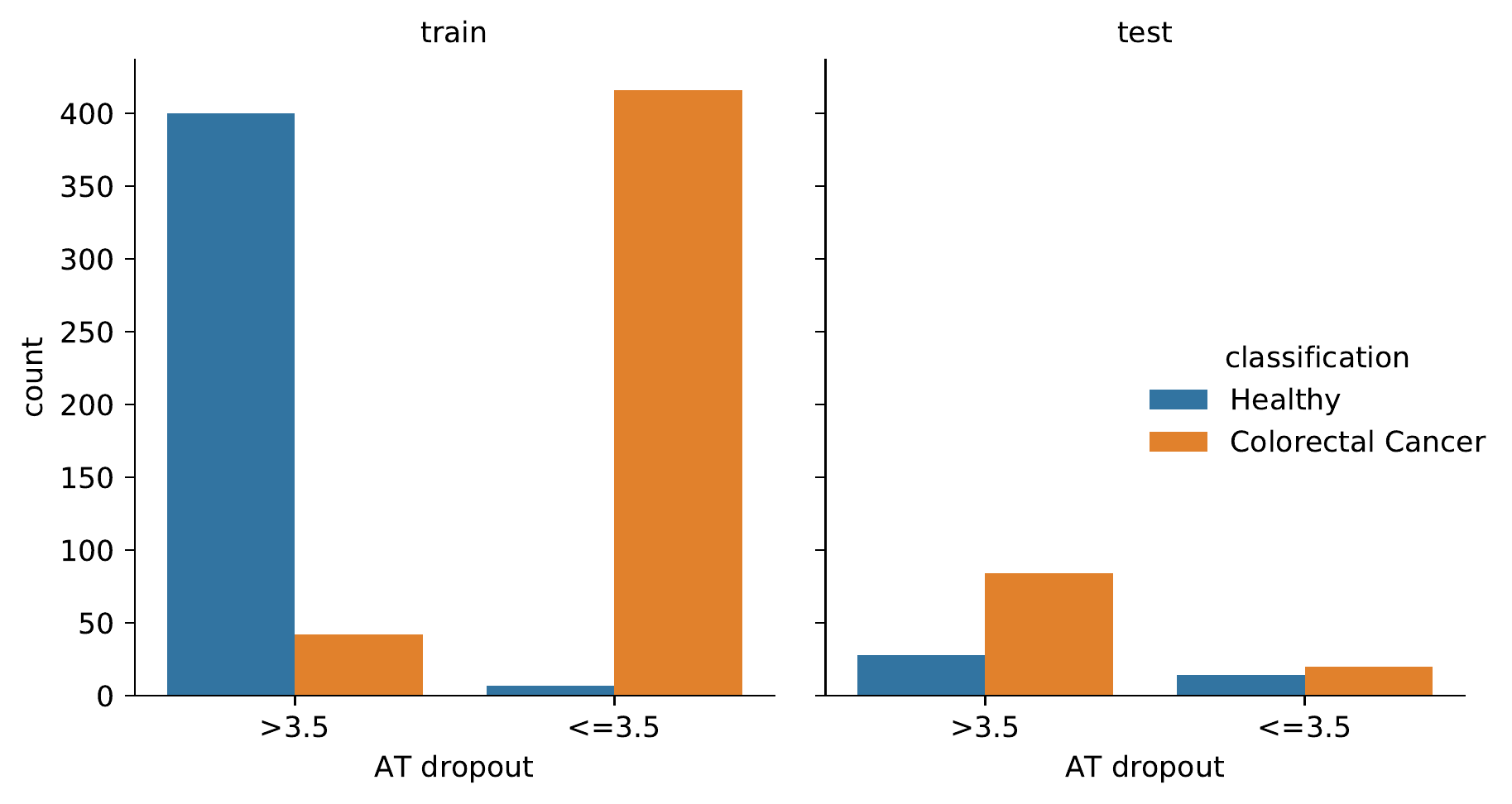}
    \caption{Sample distribution in the experiments of GC bias confounding. \label{gc_distribution}}
\end{figure}

We construct a scenario where labels in our training set are confounded by the level of GC bias. To measure GC bias, we use the Picard Tools metric of AT dropout $dropout_{AT}= \sum_{gc=0}^{50}{max(E_{gc} - O_{gc}, 0)}$, where $gc$ represents percent GC, $E_{gc} = \frac{len(R_{gc})}{len(genome)}$ (the expected number of reads overlapping the region $R_{gc}$, the union of 100 base pair windows of the genome with GC percentage $gc$, assuming uniform coverage of the genome), and $O_{gc} = \frac{\sum_{r}{I_{r \cap R_{gc}}}}{count(reads)}$, where the numerator represents the number of reads overlapping regions $R_{gc}$ \cite{Picard2018toolkit}. We focus on AT dropout because our particular sequencing procedure results in under-representation of AT-rich fragments. The sample distributions in one fold among the 5 folds are shown in Figure \ref{gc_distribution}. The confounded training set is obtained by randomly filtering healthy samples with low $dropout_{AT}$ (below 3.5) and cancer samples with high $dropout_{AT}$ (above 3.5) with probability 0.9. The sex chromosomes are removed when comparing all methods in the GC confounding experiments, since LOWESS GC correction does not perform well when chrX and chrY exist in the feature set. 

\subsection{Additional analysis results}
\begin{figure}[t!]
    \centering
        \includegraphics[width=1\linewidth]{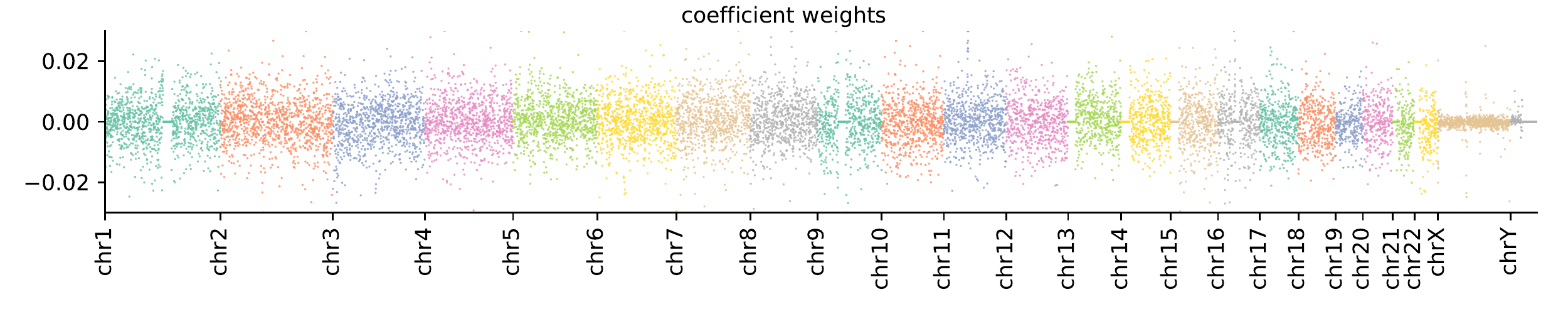}
        \includegraphics[width=1\linewidth]{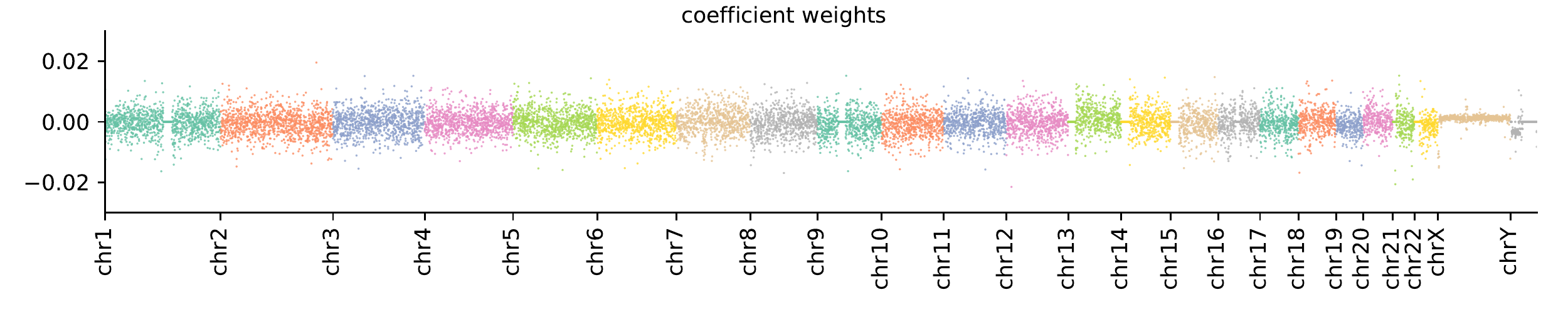}
        \includegraphics[width=0.5\linewidth]{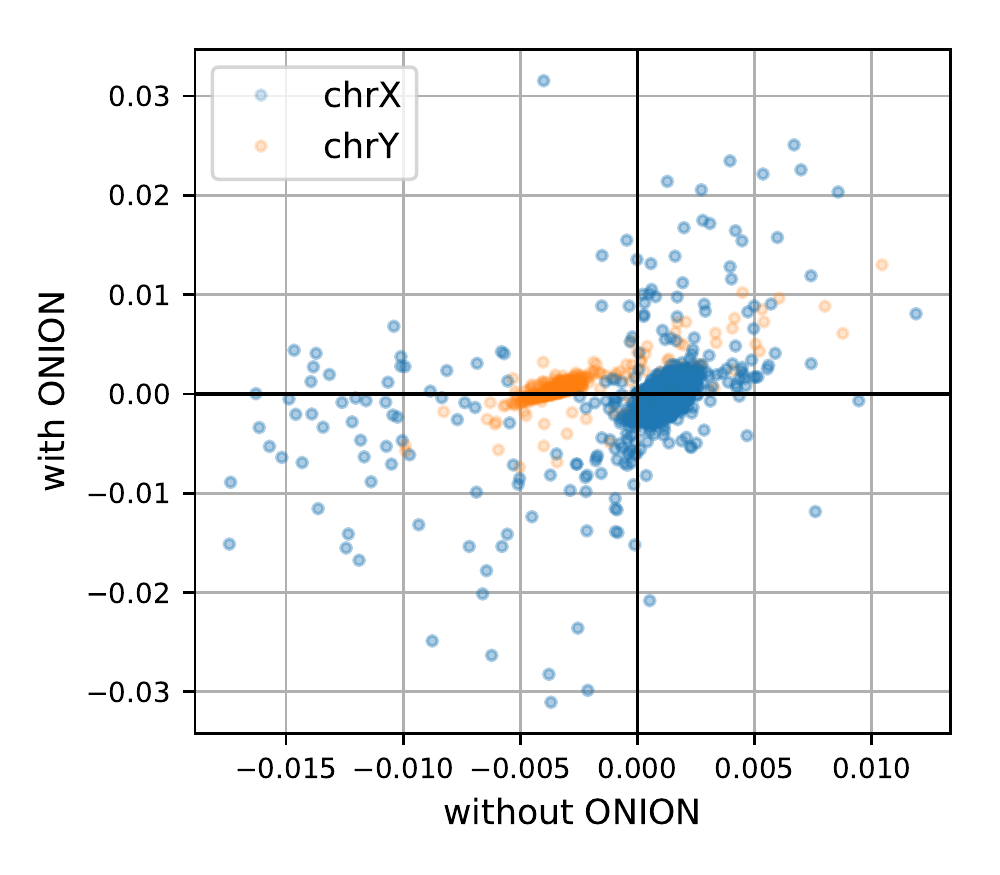}
    \caption{Classifier weights associated with each feature in the sex confounding experiment using logistic regression model with (the top panel) and without ONION (the middle panel). Note that signs of the weights associated with chrX and chrY in logreg without ONION are opposite, whereas they are all centered around zero in the model with ONION.  This is further illustrated in the bottom panel, where we plotted the weights on chrX and chrY in the logreg model with and without ONION.
    \label{onion_weights} }
\end{figure}

We examine the weights in the logistic regression model with and without ONION in the the sex confounding experiment, Table \ref{confounder-gender-gc}. Figure \ref{onion_weights} shows the weights corresponding to each feature, sorted by chromosome number. The weights associated with chrX and chrY have opposite signs in the model trained without ONION. This suggests that the model relies on sex (the confounder) to make predictions.  This pattern does not occur when using the ONION, which implies the model does not use confounders for prediction. 

\end{document}